# Enhancing Low-Resource Minority Language Translation with LLMs and Retrieval-Augmented Generation for Cultural Nuances

Chen-Chi Chang[1*], Chong-Fu Li[1], Chu-Hsuan Lee[1], and Hung-Shin Lee[2]

[1]Dept. Culture Creativity and Digital Marketing, National United University, Taiwan
[2]United Link Co., Ltd., Taiwan
`kiwi@gm.nuu.edu.tw`

**Abstract** This study investigates the challenges of translating low-resource languages by integrating Large Language Models (LLMs) with Retrieval-Augmented Generation (RAG). Various model configurations were tested on Hakka translations, with BLEU scores ranging from 12% (dictionary-only) to 31% (RAG with Gemini 2.0). The best-performing model (Model 4) combined retrieval and advanced language modeling, improving lexical coverage, particularly for specialized or culturally nuanced terms, and enhancing grammatical coherence. A two-stage method (Model 3) using dictionary outputs refined by Gemini 2.0 achieved a BLEU score of 26%, highlighting iterative correction's value and the challenges of domain-specific expressions. Static dictionary-based approaches struggled with context-sensitive content, demonstrating the limitations of relying solely on predefined resources. These results emphasize the need for curated resources, domain knowledge, and ethical collaboration with local communities, offering a framework that improves translation accuracy and fluency while supporting cultural preservation.

## 1    Introduction

Large Language Models (LLMs) serve as the transformative approach for improving translation in low-resource languages—those lacking extensive digital corpora, comprehensive linguistic datasets, and cultural representation in modern technology [1, 2]. By bridging linguistic gaps and facilitating communication within underserved communities, these methods show promise for preserving cultural heritage and fostering inclusivity in digital spaces [3]. Nevertheless, low-resource contexts present formidable challenges such as constrained training data, potential biases, and inaccuracies in translations, all of which can result in cultural misrepresentations and misunderstandings [4].

---

[*] Corresponding author

To address these challenges, Retrieval-Augmented Generation (RAG) strategies have been introduced to complement the generative capabilities of LLMs with external domain knowledge, thereby reducing hallucinations—instances in which the model generates unfounded or erroneous outputs—and enhancing overall translation precision [5,6]. However, while these innovations improve technical accuracy, they do not fully resolve the ethical and practical challenges associated with translation in crisis settings. Issues such as equitable access to translation services, the active involvement of local communities in refining and validating translations, and the need to balance technological advancements with humanitarian values remain critical concerns [7].

Given these complexities, this study examines the extent to which integrating RAG with LLMs mitigates the challenges of low-resource language translation while preserving cultural and linguistic fidelity. Accordingly, it pursues three primary objectives. First, it identifies the core constraints that hinder LLM performance in low-resource contexts. Second, it assesses the effectiveness of RAG in enhancing translation accuracy and reducing errors. Finally, it proposes a framework that integrates ethical considerations, community engagement, and technological innovations to foster sustainable, culturally informed multilingual systems. By addressing these aspects, this study seeks to bridge critical knowledge gaps, emphasizing the interplay between linguistic diversity, technological advancements, and collaboration with native speaker communities.

To provide a comprehensive analysis, this study first explores existing research on cultural nuances in minority language translation, challenges faced by low-resource languages, and the role of LLMs and RAG in addressing these issues. It then introduces the methodological framework, detailing the integration of LLMs with RAG, model design, and implementation process. The findings are presented and discussed, highlighting key improvements in translation quality and the implications of different model configurations. Finally, the study reflects on its broader contributions, acknowledging limitations and offering insights for future advancements in multilingual AI systems.

## 2 Literature Review

### 2.1 Cultural Nuances in Minority Language Translation

Cultural nuances in minority language translation present distinct challenges and possibilities, particularly in treating culturally specific words, expressions, and concepts that lack direct equivalents. Multiple studies focusing on the translation of ethnic minority terms in China draw attention to the interplay between linguistic fidelity and functional communicative goals, proposing transliteration, coinage of new terms, and hybrid strategies as means of preserving cultural authenticity while promoting comprehension [8]. These approaches emphasize the importance of balancing literal adherence to the source language with the pragmatic requirements of cross-cultural communication, highlighting the translator's role in safeguarding cultural references and traditions [9].

Similar complexities arise when translating culture-specific terms across different languages, where direct equivalents may be lacking in the target language. Strategies such as explicitation, generalization, and cultural substitution help address lexical gaps and differing cultural perspectives, aligning with the skopos of translation while maintaining cultural nuances. [10, 11]. These efforts underscore the dynamic nature of translation, where achieving cultural resonance can be just as critical as linguistic accuracy.

Translation's role extends beyond linguistic transference into a broader sociopolitical sphere, shaping national identities and addressing power asymmetries that have historically marginalized certain languages and cultures. By recognizing the role of minority languages and literature as cultural carriers, we can preserve cultural heritage, promote cultural diversity, and empower underrepresented communities by ensuring their perspectives are acknowledged and valued [12]. These studies agree that minority language translation demands nuanced strategies tailored to each cultural context. Researchers emphasize the necessity of in-depth contextual awareness, collaboration with native speakers, and a commitment to preserving cultural layers alongside achieving communicative clarity.

### 2.2 Low-Resource Languages

Low-resource languages refer to languages that suffer from a significant scarcity of linguistic resources, such as corpora, dictionaries, and annotated datasets. Neural Machine Translation (NMT) for low-resource languages has garnered increasing scholarly attention, primarily due to the scarcity of parallel corpora that hinder model training and overall translation quality. Multiple lines of inquiry have emerged to mitigate these challenges. One approach involves leveraging online and offline resources, often with custom optical character recognition (OCR) systems and automatic alignment modules, thereby expanding the pool of accessible linguistic data [13]. Researchers have also examined semantic and linguistic features to enhance translation fidelity [14] alongside techniques such as transliteration and back-translation for effective data augmentation [15]. Another effective approach involves utilizing transfer learning, where pre-trained models on high-resource language pairs are adapted to enhance translation quality in low-resource scenarios [16]. In addition, mixed training regimes and the inclusion of local dependencies through word alignments have been shown to refine further translation outputs [17], while research exploring Chinese-centric perspectives demonstrates the adaptability of NMT methodologies across diverse linguistic domains [18]. Empirical findings suggest that these interventions improve translation metrics, including BLEU scores, especially in scenarios with moderate data availability. In extreme low-resource settings, however, phrase-based statistical machine translation techniques can still outperform neural methods. Ongoing investigations continue to address emerging challenges, such as refining automated data collection processes and enhancing linguistic nuance, thereby pointing to a dynamic and evolving research landscape [19].

### 2.3 Large Language Models (LLMs)

LLMs have emerged as a significant technological advancement in natural language processing (NLP), offering remarkable capabilities for language understanding and generation across a multitude of languages. Large language models, such as ChatGPT and Gemini, have been investigated across a range of specialized domains, offering insights into their capabilities and constraints [20]. Gemini displayed higher accuracy and depth in domain-specific tasks in finance and accounting, mainly where specialized jargon and intricate calculations were involved. ChatGPT was more adept at natural language interactions and text generation, suggesting suitability for generalized tasks that demand linguistic fluency over technical precision. These observations highlight the importance of task-specific assessment when determining the most appropriate LLM for financial applications [21]. Gemini was faster in producing results and scored higher on readability, indicating an advantage in environments where rapid understanding of AI-generated content is a priority. Balanced integration of both models could be beneficial, with one providing precise recommendations and the other facilitating efficient communication and comprehension [22]. Collectively, these studies illustrate the evolving role of LLMs in diverse fields and the necessity for ongoing refinement, domain-specific customization, and rigorous validation [23]. At the same time, ChatGPT and Gemini each exhibit distinctive strengths, and further considerations regarding data governance, expert oversight, and interpretability remain central to advancing their effectiveness. The strategic adoption of these models, aligned with specific user requirements and ethical guidelines, can maximize their utility across a broad spectrum of applications.

### 2.4 Retrieval-Augmented Generation (RAG)

The effectiveness of LLMs in low-resource language scenarios presents unique challenges and opportunities. Low-resource languages, which often lack substantial digital content and comprehensive linguistic resources, can greatly benefit from innovative techniques such as RAG to enhance translation and communication. In the RAG framework, a pre-trained model with parametric memory—where knowledge is encoded within the model's parameters—is complemented by an external, non-parametric memory module. This setup employs a general-purpose fine-tuning procedure that allows the model to retrieve relevant data from external sources in real time, thereby extending its effective knowledge base and improving the quality and accuracy of the generated content [24].

One of the primary challenges in utilizing LLMs for low-resource languages is the limited availability of training data, which can lead to model biases and inadequate performance in culturally sensitive contexts. Many LLMs are trained on large, diverse datasets that may not represent the specific linguistic and cultural nuances of low-resource languages, resulting in misinterpretations and inaccuracies when generating text [25]. For example, idiomatic expressions and culturally significant terms may not be accurately translated, risking a loss of their original meaning in the generation process[26]. Additionally, the ethical considerations of deploying LLMs in culturally delicate situations highlight the need for a more nuanced approach. Addressing these challenges requires a critical examination of the interplay between data availability, model bias, and cultural relevance to ensure responsible usage of LLM technologies in low-resource contexts.

RAG represents a transformative approach in leveraging LLMs for low-resource language tasks by integrating retrieval mechanisms that enhance contextual relevance and accuracy. By dynamically accessing external knowledge sources, RAG can significantly reduce hallucinations—instances where the model generates incorrect or nonsensical information—thus improving the reliability of the outputs. This is particularly vital in fields like healthcare and finance, where accurate and timely information is crucial [27]. Moreover, RAG's ability to provide up-to-date information ensures that the generated responses reflect current knowledge and developments, which is especially important in the rapidly evolving landscape of language and technology. The adaptability of RAG systems enables them to handle various tasks, including question answering and content generation, tailored to the specific requirements of each low-resource language. Recent advancements in LLM training methodologies, such as fine-tuning and multilingual model training, further support the integration of low-resource languages into mainstream NLP applications. Techniques like few-shot learning and self-supervised training empower LLMs to generalize from limited data, allowing them to infer new vocabulary and expressions over time [28]. This adaptability is crucial for maintaining the relevance of language models as they evolve alongside the languages they represent. As the field continues to develop, ongoing research will be vital in addressing the multifaceted challenges associated with low-resource language processing and ensuring that these languages are not marginalized in the digital age. The synergy between LLMs and RAG holds promise for inclusive language technology, enabling effective communication and information access for speakers of low-resource languages.

## 3    Research Methodology

### 3.1    Integration of LLMs and RAG

In conjunction with RAG, LLMs present numerous advantages, particularly in low-resource translation. This synergistic approach enhances the accuracy and efficiency of translation processes, providing a powerful tool for bridging linguistic and cultural divides. RAG allows organizations to utilize existing corpora, culturally specific words, and knowledge bases without extensive retraining of LLMs. RAG improves the model's contextual relevance by augmenting the input with relevant retrieved data. It reduces hallucinations, which can be particularly beneficial for translating low-resource languages that often lack extensive training data. This capability significantly lowers the costs associated with developing AI systems, enabling quicker and more efficient deployment of translation applications. One of the primary benefits of RAG is its ability to enhance the outputs of LLMs by integrating information from external knowledge sources. This process allows for expanding the model's knowledge base without the complexities of retraining, thereby improving the accuracy and contextual understanding of translations. Such improvements are essential when dealing with low-resource languages, where nuances in meaning and cultural context are crucial for effective communication. By utilizing RAG, smaller organizations and businesses can compete effectively with larger counterparts without significant investments in training large models on proprietary data. This democratization of technology ensures that low-resource languages receive the attention they deserve, as companies can implement RAG-enabled applications more rapidly and at a fraction of the traditional costs. As a result, this enhances accessibility and promotes inclusivity in global discourse. The combination of RAG and fine-tuning techniques can lead to the creation of hybrid models that are both accurate and adaptable for various translation tasks. Such models can leverage the vast knowledge accessible through RAG while benefiting from the task-specific optimizations that fine-tuning provides. This hybrid methodology is particularly valuable in low-resource contexts, as it allows for the efficient allocation of computational resources while maintaining high performance in translation accuracy. In translation, addressing cultural nuances and contextual subtleties is critical. RAG-equipped LLMs can help overcome the challenges often faced in low-resource language translation by ensuring that translations preserve the original meaning and reflect the cultural context. This focus on cultural sensitivity enhances the quality of translations, fostering better understanding and appreciation across diverse ethnic groups.

### 3.2 Model Design

This study utilizes seven distinct model configurations to systematically evaluate the performance of Hakka translation in the Sìxiàn dialect (see Fig.1). Each model differs in terms of data resources, translation strategies, and post-processing mechanisms, thereby enabling a comprehensive analysis of how various approaches influence translation quality and cultural fidelity. By comparing these seven models, the study aims to discern how different combinations of dictionary-based systems, large language models, retrieval mechanisms, and iterative refinement procedures affect the overall quality, fluency, and cultural appropriateness of Sìxiàn Hakka translations. The results provide a nuanced understanding of where certain approaches—like dictionary-driven resources or real-time retrieval—excel, as well as how multi-stage or integrated pipelines can optimize translation outputs. This comparative framework thus not only advances knowledge in low-resource language processing but also sheds light on practical strategies for preserving and revitalizing culturally significant minority languages. The following subsections detail the design and rationale behind each model configuration.

Model 0 (Baseline) serves as a baseline by directly applying the Gemini 2.0 large language model to translate Mandarin text into Sìxiàn Hakka without any additional enhancement techniques. The translation process is guided by a carefully constructed prompt, but no external resources or secondary models are involved. This minimal setup aims to clarify the inherent capabilities of Gemini 2.0, allowing subsequent models to be compared against a control condition. By isolating Gemini 2.0's translation performance, the research team gains insight into the model's baseline accuracy, linguistic fluency, and potential limitations regarding culturally specific vocabulary and expressions.

Model 0a also serves as an another baseline model, directly using ChatGPT-4 to translate Mandarin text into Sìxiàn Hakka. It relies solely on a carefully designed prompt, without incorporating external resources or auxiliary models. This setup helps clarify ChatGPT-4's standalone translation performance in a low-resource language context.

Model 1 leverages a dictionary-based system offered by the Hakka AI co-creation platform, GoHakka.org, which focuses on a community-driven approach to building linguistic resources. Translations are generated through an established set of dictionary mappings and rules, emphasizing terminological consistency and transparent lexical choices. Although this approach can be less flexible than neural models in handling unstructured or creative expressions, it often excels in maintaining terminology consistency. The inclusion of a specialized dictionary grounded in Hakka community contributions underscores the importance of domain-specific knowledge when translating culturally rich content.

Model 2 incorporates GPT-4 in tandem with a RAG framework. The workflow begins with user-provided Mandarin text that undergoes segmentation (e.g., using Jieba[†], a popular toolkit for Chinese word segmentation), followed by a retrieval step that queries an external knowledge base. This knowledge base consists of context-specific Hakka entries, either curated from open resources or compiled from community input. GPT-4 then integrates these retrieved segments to generate a translation that aims to be both contextually accurate and culturally aligned. By combining GPT-4's strong generative capabilities with a non-parametric memory module (the knowledge base), the model has the potential to mitigate hallucination, maintain nuance in cultural references, and handle rare or domain-specific terms more effectively than a standalone generative approach.

Model 3 employs a sequential procedure in which Hakka output from Model 1—the dictionary-based system at GoHakka.org—undergoes refinement by Gemini 2.0. The first stage focuses on ensuring lexical consistency using established dictionary mappings, which helps address standard vocabulary usage. The second stage targets subtle linguistic elements, such as colloquial phrasing, grammatical coherence, and cultural connotation, through Gemini 2.0's advanced language modeling capabilities. This iterative process creates a feedback loop, wherein any ambiguities or errors from the dictionary-based translation can be identified and corrected, potentially yielding a more fluent and culturally appropriate final output. Observing the improvements and limitations that arise from this two-stage design offers insights into the value of multi-step, hybrid translation pipelines.

Model 3a adopts the same sequential process as Model 3, but replaces Gemini 2.0 with ChatGPT-4 for the refinement stage. ChatGPT-4 revises the Hakka output from Model 1, focusing on improving fluency, grammatical accuracy, and cultural appropriateness. This approach allows for an alternative refinement pathway, offering a complementary perspective on enhancing low-resource language translation through a hybrid process.

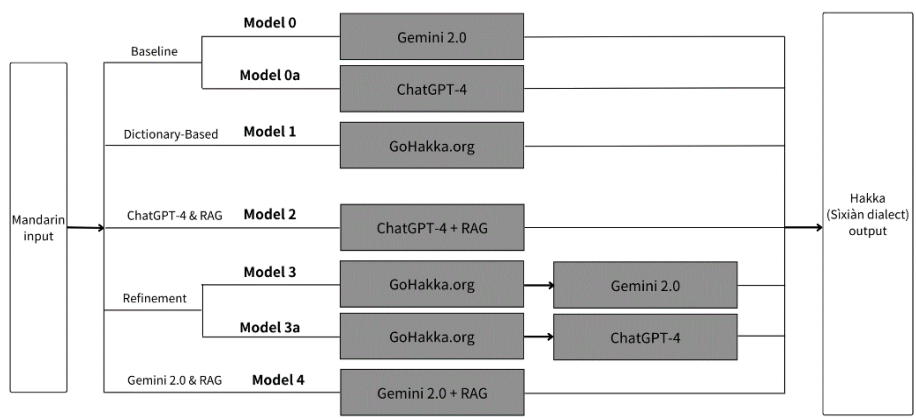

**Fig. 1.** Research flow chart.

[†] https://github.com/fxsjy/jieba

Model 4 unites Gemini 2.0 with RAG in a single pipeline, merging the strengths of deep learning and context retrieval into one cohesive system. As with Model 2, a knowledge base is first queried to extract relevant Hakka terminology and contextual information. However, instead of relying on GPT-4, Gemini 2.0 handles both the retrieval integration and the final generation step. This approach seeks to harness Gemini 2.0's language modeling capacities while supplying it with robust, context-specific references, thereby maintaining translation consistency and maximizing cultural fidelity. Model 4 is particularly designed to demonstrate how synchronous retrieval and advanced modeling can enhance performance in translating minority languages, especially when dealing with nuanced or less frequently documented terminology.

### 3.3 Implementation Phase

At the implementation stage, seven distinct models were employed to translate Mandarin input into Hakka (Sìxiàn dialect). Each model integrates a unique approach to prompt construction, dictionary-based translation, or external knowledge retrieval, thereby allowing for a thorough examination of how diverse methodological choices affect translation fluency, semantic coherence, and cultural accuracy.

During the implementation phase, Model 0 employed a straightforward prompt explicitly restricting the response length to 50 characters or fewer. The prompt is as follows: "You are an expert in the grammar of Taiwanese Hakka (Sìxiàn dialect). Please translate the user's Mandarin content into Sìxiàn Hakka word for word, in a professional and precise manner, ensuring that the sentence remains smooth and natural. Limit your response to 50 characters or fewer, and omit any additional information."

Under these constraints, Model 0 provides a direct translation from Mandarin to Hakka, aiming to maintain grammatical integrity and sentence fluency within tight length restrictions. Model 0a follows the exact same procedure, except the translation is performed by ChatGPT-4 instead of Gemini 2.0. Together, these two models establish a baseline for evaluating the inherent capabilities and limitations of Gemini 2.0 and ChatGPT-4 when no external resources or refinement steps are introduced.

Model 1 relies on the GoHakka.org platform, a community-driven and dictionary-based machine translation system, to generate Hakka text from Mandarin inputs. This approach does not use a dedicated prompt, but rather leverages the platform's built-in linguistic resources and mapping algorithms. Although it lacks the specificity of a custom prompt, Model 1 can be advantageous for ensuring consistent terminology and straightforward translations. Its dictionary-focused architecture is particularly useful for clarifying standardized vocabulary, but may require further processing to achieve nuanced, context-sensitive translations or to capture colloquial expressions that extend beyond the dictionary's scope.

Model 2 integrates GPT-4 with a RAG workflow to enhance the cultural and contextual accuracy of Hakka translations. The prompt is shown below: "You are a language expert in Taiwanese Hakka, responsible for converting the user's Mandarin sentence into a colloquial yet professional Hakka expression. First, use Jieba for tokenization, then construct a naturally fluent Hakka sentence by referring to the knowledge base entries (from Mandarin to Hakka). Ensure the output aligns with spoken usage. Retain the original punctuation, avoid adding any non-text symbols, and if a term is not found in the knowledge base, use the original Mandarin characters."

This procedure consists of three stages: (1) tokenizing the input via Jieba, (2) retrieving suitable Hakka equivalents from a specialized knowledge base, and (3) leveraging GPT-4 to integrate these terms into a coherent sentence. By fusing generative modeling with external retrieval, Model 2 reduces the risk of inaccuracies or omissions, especially for culturally significant or domain-specific expressions.

Model 3 follows a two-step translation process: it first obtains a preliminary Hakka translation from Model 1's dictionary-based mechanism and subsequently refines this output using Gemini 2.0. Model 3a follows the same procedure, except that the second-stage refinement is performed by ChatGPT-4 instead of Gemini 2.0. The prompt for the second stage is: "You are an expert in Hakka grammar. Based on your expertise, please revise the user's Hakka sentence to make it more colloquial and professional, ensuring the diction and grammar conform more closely to Hakka standards. The response is limited to 50 characters or fewer. Retain the original punctuation, refrain from inserting any non-text symbols, and provide only one sentence with colloquial modifications. Omit all additional content." By employing this two-layered approach, Model 3 and Model 3a not only retains the terminological consistency afforded by the dictionary-based translation in Model 1 but also improves grammatical accuracy, style, and cultural authenticity through post-processing by Gemini 2.0 or ChatGPT-4.

Model 4 integrates the Gemini 2.0 model with RAG in a unified procedure, combining the advantages of real-time retrieval and advanced generation. Its prompt is as follows: "You are a Hakka language expert, responsible for converting the user's Mandarin sentence into a colloquial yet professional Hakka expression. First, use Jieba to segment the text, then assemble a naturally fluent Hakka sentence according to the knowledge base entries (Mandarin column to Hakka column), ensuring it reflects spoken conventions. Retain original punctuation and avoid adding non-text symbols; if the knowledge base lacks a matching term, use the original Mandarin characters." Unlike Model 2, where GPT-4 handles the final generation, Model 4 relies on Gemini 2.0 to produce the output after the retrieval step. By merging retrieval-based enrichment with Gemini 2.0's generative capability, Model 4 endeavors to achieve higher accuracy in preserving cultural and linguistic subtleties while maintaining cohesive sentence structure. These seven models illustrate a range of methodological choices for Hakka translation, from straightforward prompt-based generation to advanced retrieval integration and iterative refinement. Each configuration addresses different aspects of lexicon consistency, grammatical rigor, and cultural alignment, thereby providing a multifaceted perspective on how modern AI techniques can be adapted to support minority language preservation and development.

## 4 Results and Discussion

The evaluation results indicate that Model 0 yields a BLEU score of 18%, Model 1 achieves 24%, Model 2 registers 21%, Model 3 attains 26%, and Model 4 stands out with 31%. Please see Table 1. Additionally, a closer examination of Model 0 and Model 3 reveals further insights into the impact of architectural enhancements. Specifically, Model 0a achieves a BLEU score of 12%, while Model 3a improves to 17%. Several factors help explain these variations in performance. Model 4 benefits from a seamless integration of RAG and Gemini 2.0, allowing the system to reference a targeted knowledge base in real time while generating outputs. This design enhances lexical coverage—especially for more specialized or culturally specific terms—and promotes grammatical accuracy through Gemini 2.0's language modeling capabilities. The retrieval step reduces the likelihood of mistranslations or omissions by anchoring the generated text in verified Hakka correspondences, and Gemini 2.0 refines syntactic structure, resulting in higher overall fluency and precision.

Model 3, which applies a two-stage process of dictionary-based translation followed by Gemini 2.0 refinement, shows an improvement over single-step system (e.g., Model 1). The iterative correction augments readability and lexical appropriateness, thus increasing the BLEU score to 26%. However, the final output remains somewhat constrained by the initial dictionary-driven text, which may not cover domain-specific phrases or exhibit the breadth of linguistic flexibility that a retrieval mechanism provides. Consequently, Model 3 does not surpass Model 4, in which both retrieval enhancement and advanced generative methods occur in a unified pipeline.

Model 1, relying solely on GoHakka.org's dictionary-based engine, achieves 24% without any custom prompts or post-processing. This score suggests that the platform's lexical mappings are reasonably comprehensive for frequently used Hakka vocabulary. Nevertheless, specialized or nuanced expressions can pose challenges, as dictionary-centric approaches typically lack the contextual adaptability inherent in large language models. Model 2, scoring 21%, benefits from GPT-4's high-level generative capabilities but may produce suboptimal translations in some instances if the retrieval process or knowledge base alignment does not fully align with real-time user input. Although Model 2 incorporates advanced language modeling, it might occasionally struggle with precise grammar or less common terminology if the retrieved data remain incomplete or the generative model prioritizes fluency over strict lexical accuracy.

The BLEU scores underscore how each model capitalizes on particular strengths or encounters limitations. Systems that combine robust retrieval methods with strong generative architectures—such as Model 4—tend to exhibit superior alignment with reference translations, leading to higher BLEU scores. Meanwhile, systems with a single-stage or dictionary-based approach can still yield usable translations but may require additional refinements or specialized data to handle the breadth of linguistic variations found in Hakka expressions.

**Table 1.** Research Results.

| Models | Workflow Design | BLEU |
|---|---|---|
| Model 0 | Baseline with Gemini 2.0 | 0.18 |
| Model 0a | Baseline with ChatGPT 4.0 | 0.12 |
| Model 1 | Dictionary-Based Machine Translation | 0.24 |
| Model 2 | GPT-4 with Retrieval-Augmented Generation | 0.21 |
| Model 3 | Dictionary-Based + Gemini 2.0 Refinement | 0.26 |
| Model 3a | Dictionary-Based + ChatGPT 4.0 Refinement | 0.17 |
| Model 4 | Integrated Gemini 2.0 + RAG | 0.31 |

# 5    Discussion and Conclusion

## 5.1    Discussion

The experimental findings underscore the value of combining robust generative capacities with targeted retrieval methods to address the inherent complexities of translating minority languages. Model 4 emerged as the strongest performer, achieving the highest BLEU score (31%) due to its seamless integration of RAG with the Gemini 2.0 model. By referencing a dedicated Hakka knowledge base in real time, it was able to incorporate specialized or culturally nuanced terms more accurately, while Gemini 2.0's advanced language modeling optimized grammaticality and overall cohesion. These results align with previous research indicating that retrieval-based mechanisms can help to mitigate "hallucinations" and contextual inconsistencies by anchoring the translation process in verified linguistic resources.

Model 3, leveraging a two-stage process (dictionary-based translation followed by Gemini 2.0 refinement), secured a BLEU score of 26%. This arrangement underscores the potential benefits of iterative correction, where an initial output—albeit limited by dictionary-driven approaches—can be refined by a more sophisticated language model that enhances both lexical choices and syntactic flow. A closer inspection further reveals that a variant of this approach, Model 3a, which enhances retrieval capabilities, achieves 17%, demonstrating the incremental gains introduced by retrieval adjustments at this stage. However, the stepwise nature of this design restricts its ability to handle domain-specific terminology or complex idiomatic expressions that fall outside the dictionary's scope. In comparison, Model 0a, which relies solely on direct dictionary lookup without any generative refinement, recorded a BLEU score of 12%, highlighting the inherent limitations of purely static retrieval approaches. Hence, while the two-stage approach surpasses simpler methods like Model 1 (24%), it does not match the integrated RAG pipeline of Model 4.

Models 1 and 2 illustrate the trade-offs between dictionary-centric strategies and single-step generation. Model 1's reliance on GoHakka.org's platform proved reasonably effective for frequently encountered Hakka vocabulary but struggled with nuanced cultural references or specialized contexts. Meanwhile, Model 2 combined GPT-4's generative strength with retrieval but achieved only a 21% BLEU score, largely due to mismatches between the knowledge base and user inputs in certain scenarios.

## 5.2    Conclusion

This outcome reinforces the necessity for well-curated resources and close alignment between retrieval mechanisms and the large language model's generative capabilities. Taken together, these results underscore several key insights: First, systems that integrate retrieval augmentation with sophisticated generation (such as Model 4) generally outperform those relying on a single approach. This synergy allows for real-time referencing of culturally specific lexicons while maintaining coherent, fluent output. Second, minority languages often carry unique idiomatic and cultural elements that can be lost without careful curation of domain-specific knowledge. Retrieval modules, comprehensive dictionaries, and refined language modeling can all play pivotal roles in preserving these subtle nuances. Third, two-stage methods (e.g., Model

3) demonstrate the benefit of refining initial outputs; however, their efficacy depends heavily on the quality of both the initial translation and the language model employed for subsequent revision. Fourth, As highlighted in earlier sections, collaboration with native speakers and local communities is essential for creating ethically responsible and contextually informed models. Transparency, consent, and fair compensation in data collection remain foundational for fostering trust and ensuring that language technology contributes to, rather than exploits, cultural preservation. Finally, the insights gleaned from Hakka translation have broader implications for other low-resource languages. By bridging linguistic gaps and facilitating communication within underserved communities, LLM-based approaches offer tangible prospects for cultural preservation and inclusivity in digital spaces. Ongoing innovation in retrieval algorithms, prompt engineering, and finetuning protocols stands to make these technologies even more adaptive and precise, thereby expanding their utility across various domains—from healthcare to legal aid—where accurate and culturally sensitive translations are paramount. In conclusion, the study's findings reinforce the transformative potential of Retrieval-Augmented Generation and advanced language modeling for low-resource language translation. Harnessing these capabilities, particularly through community-driven data curation and collaborative practices, can significantly improve the contextual accuracy, cultural fidelity, and inclusivity of AI-mediated translations. Future endeavors should further explore dynamic retrieval strategies, larger or more specialized corpora, and deeper partnerships with native speaker communities. By aligning technological innovation with cultural preservation and ethical guidelines, researchers can create multilingual systems that not only excel in translation tasks but also empower local communities and uphold the rich linguistic heritage that minority languages embody.

## 6      Limitations and Challenges

Despite the promising outcomes demonstrated across the seven experimental models—especially with regard to improved BLEU scores and linguistic fidelity—several structural and practical limitations constrain the broader applicability of LLMs in low-resource language translation.

First, ethical considerations present ongoing challenges, particularly when LLM-based approaches are deployed in underserved linguistic communities. The necessity of engaging with local speakers and cultural gatekeepers is paramount for ethically gathering and annotating data. Failure to consult these stakeholders' risks exploitation of communal knowledge and may undermine trust. Transparent and responsible data practices that protect privacy, maintain security, and uphold consent protocols are critical for ensuring that low-resource language users do not become vulnerable to misappropriation or invasive data collection.

Second, scarcity of high-quality datasets remains a central obstacle. Although dictionary-based or retrieval-augmented frameworks can mitigate some data insufficiencies, the lack of extensive, up-to-date corpora frequently leads to sampling bias. Regional idioms, dialectal variations, or newly emerging linguistic trends may be overlooked, limiting the system's effectiveness in adequately representing the diverse nuances of the target language. This shortfall is especially pressing for minority languages that exhibit significant variation in grammar, lexicon, and cultural connotations, all of which are essential for achieving precise and context-sensitive translations.

Third, biases inherent in large-scale training can distort the outputs for minority languages. LLMs typically undergo training on massive datasets that span multiple languages and cultural contexts, which may result in output that aligns with majority-language norms or fails to capture key cultural subtleties. Idiomatic expressions, metaphors, and culturally significant proverbs are often absent or underrepresented, exposing translations to the risk of semantic loss or outright misinterpretation. Contextual accuracy is thus heavily dependent on active collaboration with native speakers to refine domain-specific or culturally embedded references.

Fourth, usability limitations impede the adoption of LLM-based translation solutions in crucial domains such as healthcare or legal counseling. The accuracy and clarity of automated translations can vary widely depending on the complexity of the source text. Fixed-phrase tools may succeed in standardized scenarios but lack flexibility in nuanced interactions, whereas unconstrained machine translation can yield ambiguous outputs if the underlying model is poorly adapted to the specific domain. Ensuring that these technologies supplement, rather than replace, professional interpreters require not only context-specific fine-tuning but also comprehensive training for end-users on the models' capabilities and caveats.

Fifth, the limitations of purely quantitative metrics highlight the need for more comprehensive evaluation methods. While BLEU scores provide a valuable measure of linguistic fidelity, they cannot fully capture the cultural nuances embedded in translations. In future work, incorporating human evaluation of translations would provide a more nuanced assessment of model performance. By including real-world examples of translations, it would be possible to illustrate how cultural subtleties are either captured or missed by each model. This addition would complement the quantitative findings and provide deeper insights into the effectiveness of LLMs in preserving cultural context in minority language translations.

Finally, the "black box" nature of LLMs and the resource requirements associated with their deployment pose significant hurdles for ongoing research and operational integration. Proprietary architectures often limit transparency in how data are utilized or how specific outputs are generated. Consequently, replicating or auditing model performance becomes difficult, especially for teams with restricted budgets or computational resources. This lack of openness can stifle efforts to debug, adapt, or explain system outputs—factors that are crucial for cultivating trust and expanding the use of LLM-based tools in low-resource contexts. These challenges underscore the importance of ethical data collection, extensive collaboration with community stakeholders, and targeted model enhancements. Although advanced architectures such as retrieval-augmented generation in conjunction with powerful language models hold substantial promise, robust solutions will require a multifaceted approach—one that balances technological innovation with diligent attention to cultural, ethical, and socioeconomic realities.

## Acknowledgments

This paper is based upon work funded by the National Science and Technology Council, Taiwan under Grant No. 114-2420-H-239 -002